\documentclass[11pt,a4paper]{article}
\usepackage[hyperref]{emnlp2018}
\usepackage{times}
\usepackage{latexsym}
\usepackage{url}
\usepackage{multirow}
\usepackage{amssymb}
\usepackage{amsfonts}
\usepackage{color}
\usepackage{MnSymbol}
\usepackage{makecell}
\usepackage{arydshln}
\usepackage{graphicx}
\usepackage{microtype}
\usepackage[shortlabels]{enumitem}

\usepackage{subcaption}

\usepackage{caption} 

\usepackage{booktabs}

\makeatletter
\newcommand\footnoteref[1]{\protected@xdef\@thefnmark{\ref{#1}}\@footnotemark}
\makeatother

\newcommand{\ourl}{Multistage Fusion Process}
\newcommand{\ours}{MFP}

\newcommand{\pipelinel}{Recurrent Multistage Fusion Network}
\newcommand{\pipelines}{RMFN}

\makeatletter
   \def\tagform@#1{\maketag@@@{\normalsize(#1)\@@italiccorr}}
\makeatother

\aclfinalcopy 

\title{Multimodal Language Analysis with Recurrent Multistage Fusion}

\author{Paul Pu Liang$^1$, Ziyin Liu$^2$, Amir Zadeh$^2$, Louis-Philippe Morency$^2$ \\
$^1$Machine Learning Department, $^2$Language Technologies Institute \\
Carnegie Mellon University \\
\texttt { \{pliang,ziyinl,abagherz,morency\}@cs.cmu.edu} \\
}
\date{}

\begin{document}
\maketitle

\begin{abstract}
Computational modeling of human multimodal language is an emerging research area in natural language processing spanning the language, visual and acoustic modalities. Comprehending multimodal language requires modeling not only the interactions within each modality (intra-modal interactions) but more importantly the interactions between modalities (cross-modal interactions). In this paper, we propose the Recurrent Multistage Fusion Network (RMFN) which decomposes the fusion problem into multiple stages, each of them focused on a subset of multimodal signals for specialized, effective fusion. Cross-modal interactions are modeled using this multistage fusion approach which builds upon intermediate representations of previous stages. Temporal and intra-modal interactions are modeled by integrating our proposed fusion approach with a system of recurrent neural networks. The RMFN displays state-of-the-art performance in modeling human multimodal language across three public datasets relating to multimodal sentiment analysis, emotion recognition, and speaker traits recognition. We provide visualizations to show that each stage of fusion focuses on a different subset of multimodal signals, learning increasingly discriminative multimodal representations.
\end{abstract}

\section{Introduction}

\begin{figure}[t!]
\centering{
\includegraphics[width=\linewidth]{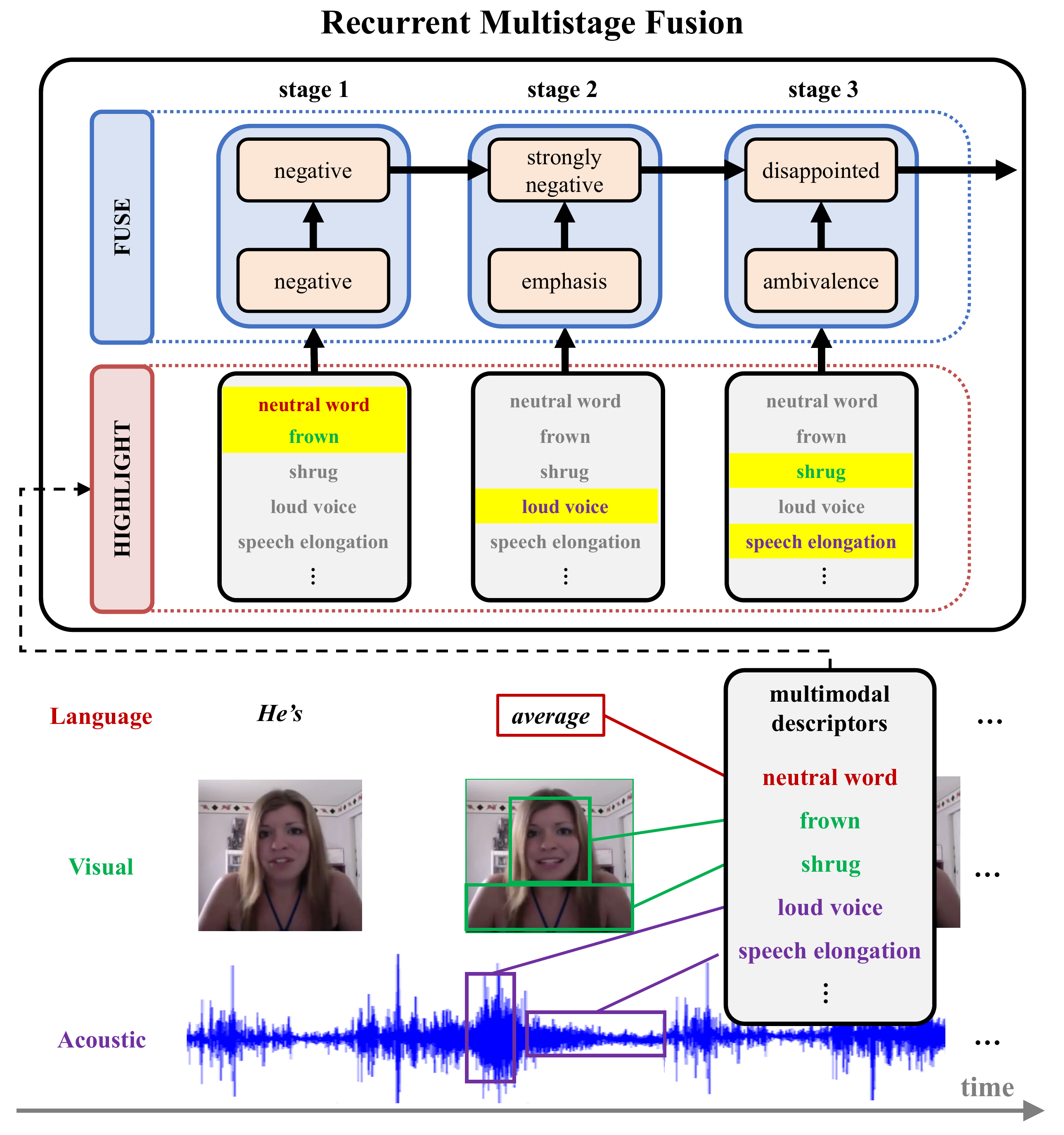}}
\caption{An illustrative example for Recurrent Multistage Fusion. At each recursive stage, a subset of multimodal signals is highlighted and then fused with previous fusion representations. The first fusion stage selects the neutral word and frowning behaviors which create an intermediate representation reflecting negative emotion when fused together. The second stage selects the loud voice behavior which is locally interpreted as emphasis before being fused with previous stages into a strongly negative representation. Finally, the third stage selects the shrugging and speech elongation behaviors that reflect ambivalence and when fused with previous stages is interpreted as a representation for the disappointed emotion. \vspace{-4mm}}
\label{fig:overview}
\end{figure}    
    

Computational modeling of human multimodal language is an upcoming research area in natural language processing. This research area focuses on modeling tasks such as multimodal sentiment analysis~\citep{morency2011towards}, emotion recognition~\citep{Busso2008IEMOCAP:Interactiveemotionaldyadic}, and personality traits recognition~\citep{Park:2014:CAP:2663204.2663260}. The multimodal temporal signals include the language (spoken words), visual (facial expressions, gestures) and acoustic modalities (prosody, vocal expressions). At its core, these multimodal signals are highly structured with two prime forms of interactions: intra-modal and cross-modal interactions~\citep{rajagopalan2016extending}. Intra-modal interactions refer to information within a specific modality, independent of other modalities. For example, the arrangement of words in a sentence according to the generative grammar of a language~\citep{reason:Chomsky57a} or the sequence of facial muscle activations for the presentation of a frown. Cross-modal interactions refer to interactions between modalities. For example, the simultaneous co-occurrence of a smile with a positive sentence or the delayed occurrence of a laughter after the end of a sentence. Modeling these interactions lie at the heart of human multimodal language analysis and has recently become a centric research direction in multimodal natural language processing~\citep{lowrank,seq2seq,Chen:2017:MSA:3136755.3136801}, multimodal speech recognition~\citep{7846320,DBLP:journals/corr/abs-1712-00489,DBLP:journals/corr/HarwathG17,DBLP:journals/corr/KamperSSL17}, as well as multimodal machine learning~\citep{factorized,NIPS2012_4683,NgiamKKNLN11}.

Recent advances in cognitive neuroscience have demonstrated the existence of multistage aggregation across human cortical networks and functions~\citep{cite-key}, particularly during the integration of multisensory information~\citep{PARISI2017208}. At later stages of cognitive processing, higher level semantic meaning is extracted from phrases, facial expressions, and tone of voice, eventually leading to the formation of higher level cross-modal concepts~\citep{PARISI2017208,cite-key}. Inspired by these discoveries, we hypothesize that the computational modeling of cross-modal interactions also requires a \textit{multistage fusion} process. In this process, cross-modal representations can build upon the representations learned during earlier stages. This decreases the burden on each stage of multimodal fusion and allows each stage of fusion to be performed in a more specialized and effective manner.


In this paper, we propose the \pipelinel \ (\pipelines) which automatically decomposes the multimodal fusion problem into multiple recursive stages. At each stage, a subset of multimodal signals is highlighted and fused with previous fusion representations (see Figure~\ref{fig:overview}). This divide-and-conquer approach decreases the burden on each fusion stage, allowing each stage to be performed in a more specialized and effective way. This is in contrast with conventional fusion approaches which usually model interactions over multimodal signals altogether in one iteration (e.g., early fusion~\citep{baltruvsaitis2017multimodal}). In \pipelines, temporal and intra-modal interactions are modeled by integrating our new multistage fusion process with a system of recurrent neural networks. Overall, \pipelines \ jointly models intra-modal and cross-modal interactions for multimodal language analysis and is differentiable end-to-end.


We evaluate \pipelines \ on three different tasks related to human multimodal language: sentiment analysis, emotion recognition, and speaker traits recognition across three public multimodal datasets. \pipelines \ achieves state-of-the-art performance in all three tasks. Through a comprehensive set of ablation experiments and visualizations, we demonstrate the advantages of explicitly defining multiple recursive stages for multimodal fusion.

\section{Related Work}
\label{Related Work}
Previous approaches in human multimodal language modeling can be categorized as follows:

\noindent \textbf{Non-temporal Models}: These models simplify the problem by using feature-summarizing temporal observations~\citep{contextmultimodalacl2017}. Each modality is represented by averaging temporal information through time, as shown for language-based sentiment analysis~\citep{Iyyer2015,DBLP:journals/corr/ChenASWC16} and multimodal sentiment analysis~\citep{abburi2016multimodal,Nojavanasghari:2016:DMF:2993148.2993176,zadeh2016multimodal,morency2011towards}. Conventional supervised learning methods are utilized to discover intra-modal and cross-modal interactions without specific model design~\cite{wang2016select,poria2016convolutional}. These approaches have trouble modeling long sequences since the average statistics do not properly capture the temporal intra-modal and cross-modal dynamics~\cite{xu2013survey}.

\noindent \textbf{Multimodal Temporal Graphical Models}: The application of graphical models in sequence modeling has been an important research problem. Hidden Markov Models (HMMs)~\cite{baum1966statistical}, Conditional Random Fields (CRFs)~\citep{Lafferty:2001:CRF:645530.655813}, and Hidden Conditional Random Fields (HCRFs)~\cite{Quattoni:2007:HCR:1313053.1313265} were shown to work well on modeling sequential data from the language~\citep{W17-4114,ma2016endtoend,journals/corr/HuangXY15} and acoustic~\citep{P2FA} modalities. These temporal graphical models have also been extended for modeling multimodal data. Several methods have been proposed including multi-view HCRFs where the potentials of the HCRF are designed to model data from multiple views~\cite{song2012multi}, multi-layered CRFs with latent variables to learn hidden spatio-temporal dynamics from multi-view data~\cite{song2012multi}, and multi-view Hierarchical Sequence Summarization models that recursively build up hierarchical representations~\citep{song2013action}.

\noindent \textbf{Multimodal Temporal Neural Networks}: More recently, with the advent of deep learning, Recurrent Neural Networks~\citep{elman1990finding,Jain:1999:RNN:553011} have been used extensively for language and speech based sequence modeling~\citep{zilly2016recurrent,DBLP:journals/corr/SoltauLS16}, sentiment analysis~\citep{Socher-etal:2013,DBLP:conf/coling/SantosG14,Glorot:2011:DAL:3104482.3104547,7435182}, and emotion recognition~\citep{speech-emotion-recognition-using-deep-neural-network-and-extreme-learning-machine,unknown,DBLP:journals/corr/abs-1803-11508}. Long-short Term Memory (LSTM) networks~\cite{hochreiter1997long} have also been extended for multimodal settings~\cite{rajagopalan2016extending} and by learning binary gating mechanisms to remove noisy modalities~\cite{Chen:2017:MSA:3136755.3136801}. Recently, more advanced models were proposed to model both intra-modal and cross-modal interactions. These use Bayesian ranking algorithms~\citep{NIPS2006_3079} to model both person-independent and person-dependent features~\citep{localglobal}, generative-discriminative objectives to learn either joint~\citep{seq2seq} or factorized multimodal representations~\citep{factorized}, external memory mechanisms to synchronize multimodal data~\citep{zadeh2018memory}, or low-rank tensors to approximate expensive tensor products~\citep{lowrank}. All these methods assume that cross-modal interactions should be discovered all at once rather than across multiple stages, where each stage solves a simpler fusion problem. Our empirical evaluations show the advantages of the multistage fusion approach.

\begin{figure*}[t!]
\centering{
\includegraphics[width=0.8\linewidth]{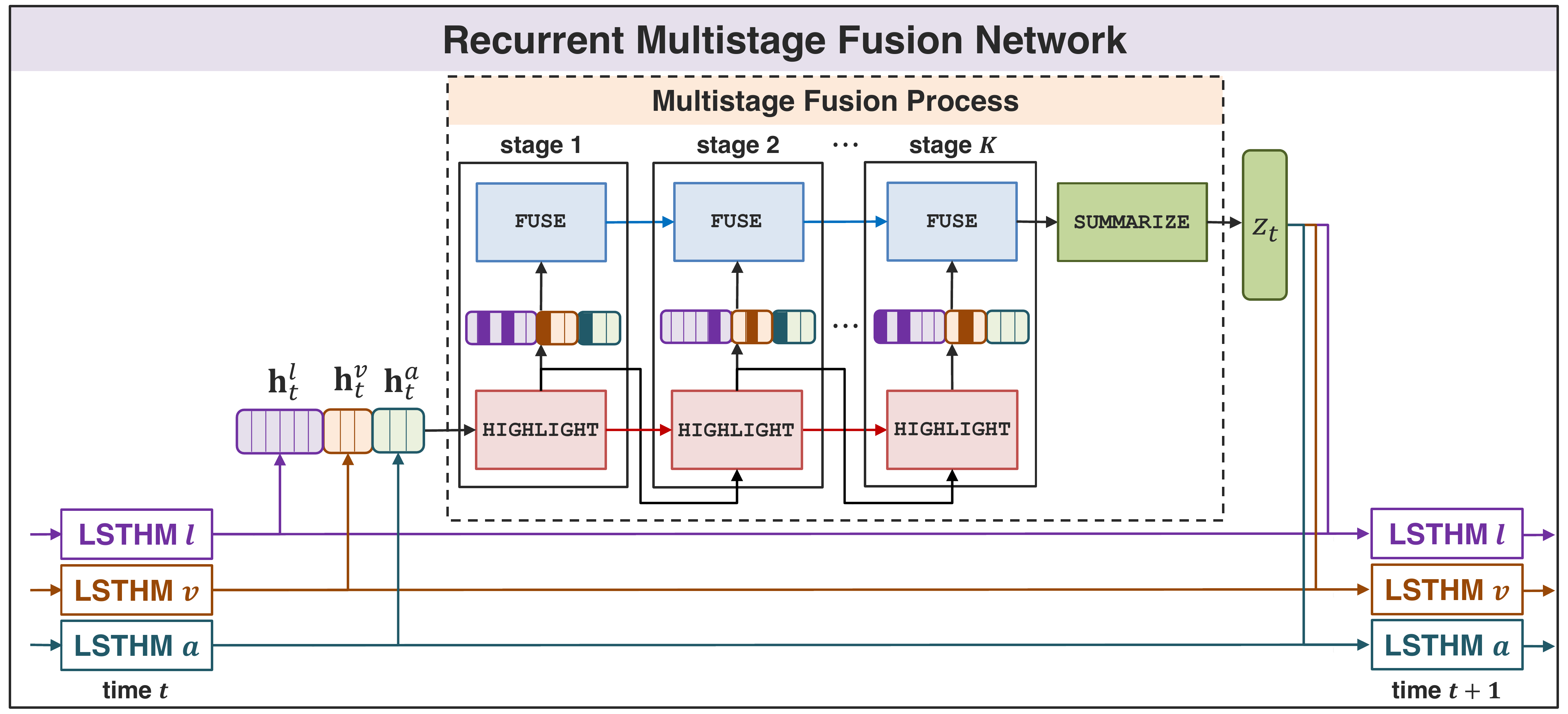}}
    \caption{The \pipelinel \ for multimodal language analysis. The \ourl \ has three modules: \texttt{HIGHLIGHT}, \texttt{FUSE} and \texttt{SUMMARIZE}. Multistage fusion begins with the concatenated intra-modal network outputs $\mathbf{h}^l_t, \mathbf{h}^v_t, \mathbf{h}^a_t$. At each stage, the \texttt{HIGHLIGHT} module identifies a subset of multimodal signals and the \texttt{FUSE} module performs local fusion before integration with previous fusion representations. The \texttt{SUMMARIZE} module translates the representation at the final stage into a cross-modal representation $\mathbf{z}_t$ to be fed back into the intra-modal recurrent networks. \vspace{-4mm}}
\label{fig:model}
\end{figure*}

\section{\pipelinel}

In this section we describe the \pipelinel \ (\pipelines) for multimodal language analysis (Figure \ref{fig:model}). Given a set of modalities $\{ l(anguage), v(isual), a(coustic) \}$, the signal from each modality $m \in \{l,v,a\}$ is represented as a temporal sequence $\mathbf{X}^m = \{\textbf{x}_{1}^m, \textbf{x}_{2}^m, \textbf{x}_{3}^m, \cdots, \textbf{x}_{T}^m \}$, where $\textbf{x}^m_{t}$ is the input at time $t$. Each sequence $\mathbf{X}^m$ is modeled with an intra-modal recurrent neural network (see subsection \ref{sec:LSTHM} for details). At time $t$, each intra-modal recurrent network will output a unimodal representation $\mathbf{h}^m_t$. The \ourl \ uses a recursive approach to fuse all unimodal representations $\mathbf{h}^m_t$ into a cross-modal representation $\mathbf{z}_t$ which is then fed back into each intra-modal recurrent network. 


\subsection{\ourl}

The \ourl \ (\ours) is a modular neural approach that performs multistage fusion to model cross-modal interactions. Multistage fusion is a divide-and-conquer approach which decreases the burden on each stage of multimodal fusion, allowing each stage to be performed in a more specialized and effective way. The \ours \ has three main modules: \texttt{HIGHLIGHT}, \texttt{FUSE} and \texttt{SUMMARIZE}. 

Two modules are repeated at each stage: \texttt{HIGHLIGHT} and \texttt{FUSE}. The \texttt{HIGHLIGHT} module identifies a subset of multimodal signals from $[\mathbf{h}^l_t,\mathbf{h}^v_t,\mathbf{h}^a_t]$ that will be used for that stage of fusion. The \texttt{FUSE} module then performs two subtasks simultaneously: a local fusion of the highlighted features and integration with representations from previous stages. Both \texttt{HIGHLIGHT} and \texttt{FUSE} modules are realized using memory-based neural networks which enable coherence between stages and storage of previously modeled cross-modal interactions. As a final step, the \texttt{SUMMARIZE} module takes the multimodal representation of the final stage and translates it into a cross-modal representation $\mathbf{z}_t$.

Figure~\ref{fig:overview} shows an illustrative example for multistage fusion. The \texttt{HIGHLIGHT} module selects ``neutral words'' and ``frowning'' expression for the first stage. The local and integrated fusion at this stage creates a representation reflecting negative emotion. For stage 2, the \texttt{HIGHLIGHT} module identifies the acoustic feature ``loud voice''. The local fusion at this stage interprets it as an expression of emphasis and is fused with the previous fusion results to represent a strong negative emotion. Finally, the highlighted features of ``shrug'' and ``speech elongation'' are selected and are locally interpreted as ``ambivalence''. The integration with previous stages then gives a representation closer to ``disappointed''. 


\subsection{Module Descriptions}
\label{multistage}

In this section, we present the details of the three multistage fusion modules: \texttt{HIGHLIGHT}, \texttt{FUSE} and \texttt{SUMMARIZE}. Multistage fusion begins with the concatenation of intra-modal network outputs $\mathbf{h}_t = \bigoplus_{m \in M} \mathbf{h}^m_t$. We use superscript $^{[k]}$ to denote the indices of each stage $k = 1, \cdots, K$ during $K$ total stages of multistage fusion. Let $\Theta$ denote the neural network parameters across all modules. 

\texttt{HIGHLIGHT}: At each stage $k$, a subset of the multimodal signals represented in $\mathbf{h}_t$ will be automatically highlighted for fusion. Formally, this module is defined by the process function $f_{H}$: 
\begin{equation}\label{eq:highlight}
\mathbf{a}_t^{[k]} = f_{H} (\mathbf{h}_t \ ; \ \mathbf{a}_t^{[1:k-1]}, \Theta)
\end{equation}
where at stage $k$, $\mathbf{a}_t^{[k]}$ is a set of attention weights which are inferred based on the previously assigned attention weights $\mathbf{a}_t^{[1:k-1]}$. As a result, the highlights at a specific stage $k$ will be dependent on previous highlights. To fully encapsulate these dependencies, the attention assignment process is performed in a recurrent manner using a LSTM which we call the \texttt{HIGHLIGHT} LSTM. The initial \texttt{HIGHLIGHT} LSTM memory at stage 0, $\mathbf{c}_t^{\texttt{HIGHLIGHT}[0]}$, is initialized using a network $\mathcal{M}$ that maps $\mathbf{h}_t$ into LSTM memory space:
\begin{equation}
\mathbf{c}_t^{\texttt{HIGHLIGHT}[0]} = \mathcal{M} (\mathbf{h}_t \ ; \ \Theta)
\end{equation}
This allows the memory mechanism of the \texttt{HIGHLIGHT} LSTM to dynamically adjust to the intra-modal representations $\mathbf{h}_t$. The output of the \texttt{HIGHLIGHT} LSTM $\mathbf{h}_t^{\texttt{HIGHLIGHT}[k]}$ is softmax activated to produce attention weights $\mathbf{a}_t^{[k]}$ at every stage $k$ of the multistage fusion process:
\begin{equation}
{\mathbf{a}_t^{[k]}}_j = \frac{\exp \ ( {\mathbf{h}_t^{\texttt{HIGHLIGHT}[k]}}_j) }{\sum_{d=1}^{|\mathbf{h}_t^{\texttt{HIGHLIGHT}[k]}|} \exp \ ( {\mathbf{h}_t^{\texttt{HIGHLIGHT}[k]}}_d) } 
\end{equation}
and ${\mathbf{a}_t^{[k]}}$ is fed as input into the \texttt{HIGHLIGHT} LSTM at stage $k+1$. Therefore, the \texttt{HIGHLIGHT} LSTM functions as a decoder LSTM~\citep{DBLP:journals/corr/SutskeverVL14,DBLP:journals/corr/ChoMGBSB14} in order to capture the dependencies on previous attention assignments. Highlighting is performed by element-wise multiplying the attention weights $\mathbf{a}_t^{[k]}$ with the concatenated intra-modal representations $\mathbf{h}_t$:
\begin{equation}\label{eq:discover}
\tilde{\mathbf{h}}_t^{[k]} = \mathbf{h}_t \odot \mathbf{a}_t^{[k]}
\end{equation}
where $\odot$ denotes the Hadamard product and $\tilde{\mathbf{h}}_t^{[k]}$ are the attended multimodal signals that will be used for the fusion at stage $k$.

\texttt{FUSE}: The highlighted multimodal signals are simultaneously fused in a local fusion and then integrated with fusion representations from previous stages. Formally, this module is defined by the process function $f_{F}$:
\begin{equation}\label{eq:fuse}
\mathbf{s}_t^{[k]} = f_{F} (\tilde{\mathbf{h}}_t^{[k]} \ ; \ \mathbf{s}_t^{[1:k-1]}, \Theta)
\end{equation}
where $\mathbf{s}_t^{[k]}$ denotes the integrated fusion representations at stage $k$. We employ a \texttt{FUSE} LSTM to simultaneously perform the local fusion and the integration with previous fusion representations. The \texttt{FUSE} LSTM input gate enables a local fusion while the \texttt{FUSE} LSTM forget and output gates enable integration with previous fusion results. The initial \texttt{FUSE} LSTM memory at stage 0, $\mathbf{c}_t^{\texttt{FUSE}[0]}$, is initialized using random orthogonal matrices~\citep{DBLP:journals/corr/ArjovskySB15,DBLP:journals/corr/LeJH15}.

\texttt{SUMMARIZE}: After completing $K$ recursive stages of \texttt{HIGHLIGHT} and \texttt{FUSE}, the \texttt{SUMMARIZE} operation generates a cross-modal representation using all final fusion representations $\mathbf{s}_t^{[1:K]}$.
Formally, this operation is defined as:
\begin{equation}
\mathbf{z}_t = \mathcal{S}(\mathbf{s}_t^{[1:K]} \ ; \ \Theta)
\end{equation}
where $\mathbf{z}_t$ is the final output of the multistage fusion process and represents all cross-modal interactions discovered at time $t$. The summarized cross-modal representation is then fed into the intra-modal recurrent networks as described in the subsection~\ref{sec:LSTHM}.

\subsection{System of Long Short-term Hybrid Memories}
\label{sec:LSTHM}

To integrate the cross-modal representations $\mathbf{z}_{t}$ with the temporal intra-modal representations, we employ a system of Long Short-term Hybrid Memories (LSTHMs)~\citep{zadeh2018multi}. The LSTHM extends the LSTM formulation to include the cross-modal representation $\mathbf{z}_{t}$ in a hybrid memory component:
\par\nobreak
\vspace{-4mm}
{\small
\begin{align}
\mathbf{i}_{t+1}^m &= \sigma(\mathbf{W}_i^m\ \mathbf{x}^m_{t+1}+\mathbf{U}^m_i\ \mathbf{h}^m_{t}+\mathbf{V}^m_i\ \mathbf{z}_{t}+\mathbf{b}^m_{i}) \\
\mathbf{f}^m_{t+1} &= \sigma(\mathbf{W}^m_{f}\ \mathbf{x}^m_{t+1} + \mathbf{U}^m_{f}\ \mathbf{h}^m_{t} + \mathbf{V}^m_f\ \mathbf{z}_{t}+\mathbf{b}^m_{f}) \\
\mathbf{o}^m_{t+1} &= \sigma(\mathbf{W}^m_{o}\ \mathbf{x}^m_{t+1} + \mathbf{U}^m_{o}\ \mathbf{h}^m_{t} + \mathbf{V}^m_o\ \mathbf{z}_{t}+\mathbf{b}^m_{o}) \\
\bar{\mathbf{c}}_{t+1}^m &= \mathbf{W}_{\bar{c}}^m\ \mathbf{x}^m_{t+1} + \mathbf{U}_{\bar{c}}^m\ \mathbf{h}^m_{t} + \mathbf{V}_{\bar{c}}^m\ \mathbf{z}_{t} + \mathbf{b}^m_{\bar{c}} \\
\mathbf{c}^m_{t+1} &= \mathbf{f}^m_t \odot \mathbf{\mathbf{c}}^m_{t} + \mathbf{i}^m_t \odot tanh(\bar{\mathbf{c}}_{t+1}^m) \\
\mathbf{h}^m_{t+1} &= \mathbf{o}^m_{t+1} \odot tanh(\mathbf{c}^m_{t+1})
\end{align}
}%
where $\sigma$ is the (hard-)sigmoid activation function, $tanh$ is the tangent hyperbolic activation function, $\odot$ denotes the Hadamard product. $\mathbf{i},\mathbf{f}$ and $\mathbf{o}$ are the input, forget and output gates respectively. $\bar{\mathbf{c}}_{t+1}^m$ is the proposed update to the hybrid memory $\mathbf{c}^m_t$ at time $t+1$ and $\mathbf{h}^m_t$ is the time distributed output of each modality. The cross-modal representation $\mathbf{z}_t$ is modeled by the \ourl \ as discussed in subsection~\ref{multistage}. The hybrid memory $\mathbf{c}^m_t$ contains both intra-modal interactions from individual modalities $\mathbf{x}^m_t$ as well as the cross-modal interactions captured in $\mathbf{z}_t$.

\subsection{Optimization}
The multimodal prediction task is performed using a final representation $\mathcal{E}$ which integrate (1) the last outputs from the LSTHMs and (2) the last cross-modal representation $\mathbf{z}_T$. Formally, $\mathcal{E}$ is defined as: 
\begin{equation}
\mathcal{E} = (\bigoplus_{m \in M} \mathbf{h}^m_T) \bigoplus \mathbf{z}_T
\end{equation}
where $\bigoplus$ denotes vector concatenation. $\mathcal{E}$ can then be used as a multimodal representation for supervised or unsupervised analysis of multimodal language. It summarizes all modeled intra-modal and cross-modal representations from the multimodal sequences. \pipelines \ is differentiable end-to-end which allows the network parameters $\Theta$ to be learned using gradient descent approaches.

\section{Experimental Setup}

To evaluate the performance and generalization of \pipelines, three domains of human multimodal language were selected: multimodal sentiment analysis, emotion recognition, and speaker traits recognition. 

\subsection{Datasets}

All datasets consist of monologue videos. The speaker's intentions are conveyed through three modalities: language, visual and acoustic.

\noindent \textbf{Multimodal Sentiment Analysis} involves analyzing speaker sentiment based on video content. Multimodal sentiment analysis extends conventional language-based sentiment analysis to a multimodal setup where both verbal and non-verbal signals contribute to the expression of sentiment. We use \textbf{CMU-MOSI}~\cite{zadeh2016multimodal} which consists of 2199 opinion segments from online videos each annotated with sentiment in the range [-3,3]. 

\noindent \textbf{Multimodal Emotion Recognition} involves identifying speaker emotions based on both verbal and nonverbal behaviors. We perform experiments on the \textbf{IEMOCAP} dataset~\cite{Busso2008IEMOCAP:Interactiveemotionaldyadic} which consists of 7318 segments of recorded dyadic dialogues annotated for the presence of human emotions happiness, sadness, anger and neutral.

\noindent \textbf{Multimodal Speaker Traits Recognition} involves recognizing speaker traits based on multimodal communicative behaviors. \textbf{POM}~\cite{Park:2014:CAP:2663204.2663260} contains 903 movie review videos each annotated for 12 speaker traits: confident (con), passionate (pas), voice pleasant (voi), credible (cre), vivid (viv), expertise (exp), reserved (res), trusting (tru), relaxed (rel), thorough (tho), nervous (ner), persuasive (per) and humorous (hum). 

\subsection{Multimodal Features and Alignment}
GloVe word embeddings~\cite{pennington2014glove}, Facet~\cite{emotient} and COVAREP~\cite{degottex2014covarep} are extracted for the language, visual and acoustic modalities respectively \footnote{Details on feature extraction are in supplementary.}. Forced alignment is performed using P2FA~\cite{P2FA} to obtain the exact utterance times of each word. We obtain the aligned video and audio features by computing the expectation of their modality feature values over each word utterance time interval~\cite{factorized}.




\subsection{Baseline Models}
\label{sec:base}

\newcolumntype{K}[1]{>{\centering\arraybackslash}p{#1}}
\definecolor{gg}{RGB}{45,190,45}

\begin{table}[t!]
\fontsize{7.5}{10}\selectfont
\setlength\tabcolsep{1.3pt}
\begin{tabular}{l : *{16}{K{1.22cm}}}
\Xhline{3\arrayrulewidth}
Dataset & \multicolumn{5}{c}{\textbf{CMU-MOSI}} \\
Task & \multicolumn{5}{c}{Sentiment} \\
Metric       & A$2$ $\uparrow $ & F1 $\uparrow $ & A$7$ $\uparrow $ & MAE $\downarrow $ & Corr $\uparrow $\\ 
\Xhline{0.5\arrayrulewidth}
SOTA3 & 76.5 & 74.5 & 33.2 & 0.968 & 0.622  \\
SOTA2 & 77.1	& 77.0 & 34.1 & 0.965  & 0.625  \\
SOTA1 & {77.4}	 & {77.3}   & {34.7} 	&  {0.955}    	& {0.632} \\
\Xhline{0.5\arrayrulewidth}
{\pipelines}      		& \textbf{78.4} & \textbf{78.0} & \textbf{38.3} & \textbf{0.922} & \textbf{0.681} \\
\Xhline{0.5\arrayrulewidth}
$\Delta_{SOTA}$	& \textcolor{gg}{$\uparrow $ \textbf{1.0}} & \textcolor{gg}{$\uparrow $ \textbf{0.7}} & \textcolor{gg}{$\uparrow $ \textbf{3.6}} & \textcolor{gg}{$\downarrow $ \textbf{0.033}} & \textcolor{gg}{$\uparrow $ \textbf{0.049}} \\ 
\Xhline{3\arrayrulewidth}
\end{tabular}
\caption{Sentiment prediction results on CMU-MOSI. Best results are highlighted in bold and $\Delta_{SOTA}$ shows improvement over previous state of the art (SOTA). The \pipelines \ significantly outperforms the current SOTA across all evaluation metrics. Improvements highlighted in green. \vspace{-4mm}}
\label{mosi}
\end{table}
\definecolor{gg}{RGB}{45,190,45}

\begin{table}[tb]
\fontsize{7.5}{10}\selectfont
\setlength\tabcolsep{1.3pt}
\begin{tabular}{l : *{16}{K{0.73cm}}}
\Xhline{3\arrayrulewidth}
Dataset & \multicolumn{8}{c}{\textbf{IEMOCAP Emotions}} \\
Task & \multicolumn{2}{c}{Happy} & \multicolumn{2}{c}{Sad} & \multicolumn{2}{c}{Angry} & \multicolumn{2}{c}{Neutral} \\
Metric        	& A$2$ $\uparrow $ & F1 $\uparrow $ & A$2$ $\uparrow $ & F1 $\uparrow $ & A$2$ $\uparrow $ & F1 $\uparrow $ & A$2$ $\uparrow $ & F1 $\uparrow $ \\ 
\Xhline{0.5\arrayrulewidth}
SOTA3 & 86.1 & 83.6 & 83.2 & 81.7 & 85.0 & 84.2 & 68.5 & 66.7\\
SOTA2 & 86.5 & 84.0 & 83.4 & 82.1 & 85.1 & 84.3 & 68.8 & 68.5 \\
SOTA1 & 86.7 & 84.2 & 83.5 & 82.8 & \textbf{85.2} & 84.5 & \textbf{69.6} & \textbf{69.2} \\
\Xhline{0.5\arrayrulewidth}
{\pipelines}      		& \textbf{87.5} & \textbf{85.8} & \textbf{83.8} & \textbf{82.9} & {85.1} & \textbf{84.6} & {69.5} & {69.1} \\ 
\Xhline{0.5\arrayrulewidth}
$\Delta_{SOTA}$	& \textcolor{gg}{$\uparrow $ \textbf{0.8}} & \textcolor{gg}{$\uparrow $ \textbf{1.6}} & \textcolor{gg}{$\uparrow $ \textbf{0.3}} & \textcolor{gg}{$\uparrow $ \textbf{0.1}} & -- & \textcolor{gg}{$\uparrow $ \textbf{0.1}} & -- & -- \\ 
\Xhline{3\arrayrulewidth}
\end{tabular}
\caption{Emotion recognition results on IEMOCAP. Best results are highlighted in bold and $\Delta_{SOTA}$ shows improvement over previous SOTA. The \pipelines \ significantly outperforms the current SOTA across evaluation metrics except $\Delta_{SOTA}$ entries in gray. Improvements highlighted in green. \vspace{-4mm}}
\label{iemocap}
\end{table}
\newcolumntype{K}[1]{>{\centering\arraybackslash}p{#1}}
\definecolor{gg}{RGB}{45,190,45}

\begin{table*}[!htbp]
\fontsize{7}{10}\selectfont
\setlength\tabcolsep{1.3pt}
\begin{tabular}{l : *{16}{K{1.15cm}}}
\Xhline{3\arrayrulewidth}
Dataset & \multicolumn{16}{c}{\textbf{POM Speaker Personality Traits}} \\
Task & \multicolumn{1}{c}{Con} & \multicolumn{1}{c}{Pas} & \multicolumn{1}{c}{Voi} & \multicolumn{1}{c}{Cre} & \multicolumn{1}{c}{Viv} & \multicolumn{1}{c}{Exp} & \multicolumn{1}{c}{Res} & \multicolumn{1}{c}{Rel} & \multicolumn{1}{c}{Tho} & \multicolumn{1}{c}{Ner} & \multicolumn{1}{c}{Per} & \multicolumn{1}{c}{Hum}\\
Metric & A$7$ $\uparrow $ & A$7$ $\uparrow $ & A$7$ $\uparrow $ & A$7$ $\uparrow $ & A$7$ $\uparrow $ & A$7$ $\uparrow $ & A$5$ $\uparrow $ & A$5$ $\uparrow $ & A$5$ $\uparrow $ & A$5$ $\uparrow $ & A$7$ $\uparrow $ & A$6$ $\uparrow $ \\ 
\Xhline{3\arrayrulewidth}
SOTA3 & 26.6&31.0&33.0&29.6&32.5&30.5&34.0&50.7&42.9&44.8&28.1&40.4\\
SOTA2 & 29.1&33.0&34.0&31.5&36.5&31.0&36.9&52.2&45.8&47.3&31.0&44.8 \\
SOTA1 & {34.5}&{35.5}&\textbf{37.4}&{34.5}&{36.9}&{36.0}&{38.4}&{53.2}&{47.3}&{47.8}&{34.0}&\textbf{47.3} \\
\Xhline{0.5\arrayrulewidth}
{\pipelines }  &  \textbf{37.4} & \textbf{38.4} &\textbf{37.4} &\textbf{37.4} &\textbf{38.9} &\textbf{38.9} &\textbf{39.4} &\textbf{53.7} &\textbf{48.3} &\textbf{48.3} &\textbf{35.0} &{46.8}\\
\Xhline{0.5\arrayrulewidth}
$\Delta_{SOTA}$	& \textcolor{gg}{$\uparrow $ \textbf{2.9}} & \textcolor{gg}{$\uparrow $ \textbf{2.9}} & \textcolor{gg}{\textbf{0.0}} & \textcolor{gg}{$\uparrow $ \textbf{2.9}} & \textcolor{gg}{$\uparrow $ \textbf{2.0}} & \textcolor{gg}{$\uparrow $ \textbf{3.9}} & \textcolor{gg}{$\uparrow $ \textbf{1.0}} & \textcolor{gg}{$\uparrow $ \textbf{0.5}} & \textcolor{gg}{$\uparrow $ \textbf{1.0}} & \textcolor{gg}{$\uparrow $ \textbf{0.5}} & \textcolor{gg}{$\uparrow $ \textbf{1.0}} & --\\ 
\Xhline{3\arrayrulewidth}
\end{tabular}
\caption{Results for personality trait recognition on POM. Best results are highlighted in bold and $\Delta_{SOTA}$ shows improvement over previous SOTA. The \ours \ significantly outperforms the current SOTA across all evaluation metrics except the $\Delta_{SOTA}$ entries highlighted in gray. Improvements are highlighted in green. \vspace{-4mm}}
\label{pom}
\end{table*}

We compare to the following models for multimodal machine learning: \textbf{MFN} \citep{zadeh2018memory} synchronizes multimodal sequences using a multi-view gated memory. It is the current state of the art on CMU-MOSI and POM. \textbf{MARN} \citep{zadeh2018multi} models intra-modal and cross-modal interactions using multiple attention coefficients and hybrid LSTM memory components. \textbf{GME-LSTM(A)} \citep{Chen:2017:MSA:3136755.3136801} learns binary gating mechanisms to remove noisy modalities that are contradictory or redundant for prediction. \textbf{TFN} \cite{tensoremnlp17} models unimodal, bimodal and trimodal interactions using tensor products. \textbf{BC-LSTM} \cite{contextmultimodalacl2017} performs context-dependent sentiment analysis and emotion recognition, currently state of the art on IEMOCAP. \textbf{EF-LSTM} concatenates the multimodal inputs and uses that as input to a single LSTM \citep{Hochreiter:1997:LSM:1246443.1246450}. We also implement the Stacked, (\textbf{EF-SLSTM}) \citep{6638947} Bidirectional (\textbf{EF-BLSTM}) \citep{Schuster:1997:BRN:2198065.2205129} and Stacked Bidirectional (\textbf{EF-SBLSTM}) LSTMs. 
For descriptions of the remaining baselines, we refer the reader to \textbf{EF-HCRF} \cite{Quattoni:2007:HCR:1313053.1313265}, \textbf{EF/MV-LDHCRF} \cite{morency2007latent}, \textbf{MV-HCRF} \cite{song2012multi}, \textbf{EF/MV-HSSHCRF} \cite{song2013action}, \textbf{MV-LSTM} \cite{rajagopalan2016extending}, \textbf{DF} \cite{Nojavanasghari:2016:DMF:2993148.2993176}, \textbf{SAL-CNN} \cite{wang2016select}, \textbf{C-MKL} \cite{poria2015deep}, \textbf{THMM} \cite{morency2011towards}, \textbf{SVM} \cite{cortes1995support,Park:2014:CAP:2663204.2663260} and \textbf{RF} \cite{Breiman:2001:RF:570181.570182}. 


\subsection{Evaluation Metrics}
For classification, we report accuracy A$c$ where $c$ denotes the number of classes and F1 score. For regression, we report Mean Absolute Error MAE and Pearson's correlation $r$. For MAE lower values indicate stronger performance. For all remaining metrics, higher values indicate stronger performance.
\section{Results and Discussion}

\subsection{Performance on Multimodal Language}
\label{results}
Results on CMU-MOSI, IEMOCAP and POM are presented in Tables~\ref{mosi},~\ref{iemocap} and~\ref{pom} respectively\footnote{Results for all individual baseline models are in supplementary. State-of-the-art (SOTA)1/2/3 represent the three best performing baseline models on each dataset.}. We achieve state-of-the-art or competitive results for all domains, highlighting \pipelines's capability in human multimodal language analysis. We observe that \pipelines \ does not improve results on IEMOCAP neutral emotion and the model outperforming RMFN is a memory-based fusion baseline~\citep{zadeh2018memory}. We believe that this is because neutral expressions are quite idiosyncratic. Some people may always look angry given their facial configuration (e.g., natural eyebrow raises of actor Jack Nicholson). In these situations, it becomes useful to compare the current image with a memorized or aggregated representation of the speaker's face. Our proposed multistage fusion approach can easily be extended to memory-based fusion methods.





\subsection{Analysis of Multistage Fusion}
\newcolumntype{K}[1]{>{\centering\arraybackslash}p{#1}}
\definecolor{gg}{RGB}{45,190,45}

\begin{table}[t!]
\fontsize{7.5}{10}\selectfont
\setlength\tabcolsep{2.0pt}
\begin{tabular}{l : *{16}{K{0.98cm}}}
\Xhline{3\arrayrulewidth}
Dataset & \multicolumn{5}{c}{\textbf{CMU-MOSI}} \\
Task & \multicolumn{5}{c}{Sentiment} \\
Metric       & A$2$ $\uparrow $ & F1 $\uparrow $ & A$7$ $\uparrow $ & MAE $\downarrow $ & Corr $\uparrow $\\ 
\Xhline{0.5\arrayrulewidth}
\pipelines-R1 \ \ \ \ \ \ \ \ \ \ \	& {75.5} & {75.5} & {35.1} & {0.997} & {0.653} \\
\pipelines-R2 \ \ \ \ \ \ \ \ \ \ \	& {76.4} & {76.4} & {34.5} & {0.967} & {0.642} \\
\pipelines-R3 \ \ \ \ \ \ \ \ \ \ \	& \textbf{78.4} & \textbf{78.0} & \textbf{38.3} & \textbf{0.922} & \textbf{0.681} \\
\pipelines-R4 \ \ \ \ \ \ \ \ \ \ \	& {76.0} & {76.0} & {36.0} & {0.999} & {0.640} \\
\pipelines-R5 \ \ \ \ \ \ \ \ \ \ \	& {75.5} & {75.5} & {30.9} & {1.009} & {0.617} \\
\pipelines-R6 \ \ \ \ \ \ \ \ \ \ \	& {70.4} & {70.5} & {30.8} & {1.109} & {0.560} \\
\Xhline{0.5\arrayrulewidth}
\pipelines	& \textbf{78.4} & \textbf{78.0} & \textbf{38.3} & \textbf{0.922} & \textbf{0.681} \\
\Xhline{3\arrayrulewidth}
\end{tabular}
\caption{Effect of varying the number of stages on CMU-MOSI sentiment analysis performance. Multistage fusion improves performance as compared to single stage fusion. \vspace{-4mm}}
\label{length}
\end{table}
\newcolumntype{K}[1]{>{\centering\arraybackslash}p{#1}}
\definecolor{gg}{RGB}{45,190,45}

\begin{table}[t!]
\fontsize{7.5}{10}\selectfont
\setlength\tabcolsep{2.0pt}
\begin{tabular}{l : *{16}{K{0.8cm}}}
\Xhline{3\arrayrulewidth}
Dataset & \multicolumn{5}{c}{\textbf{CMU-MOSI}} \\
Task & \multicolumn{5}{c}{Sentiment} \\
Metric       & A$2$ $\uparrow $ & F1 $\uparrow $ & A$7$ $\uparrow $ & MAE $\downarrow $ & Corr $\uparrow $\\ 
\Xhline{0.5\arrayrulewidth}
MARN	& 77.1	& 77.0 & 34.7 & 0.968  & 0.625 \\
{\pipelines } (no \ours)  & {76.5} & {76.5} &  30.8	&  {0.998} & 0.582 \\ 
\pipelines \ (no \texttt{HIGHLIGHT})   & {77.9} & {77.9} & {35.9} & {0.952} & {0.666} \\
\Xhline{0.5\arrayrulewidth}
{\pipelines}      		& \textbf{78.4} & \textbf{78.0} & \textbf{38.3} & \textbf{0.922} & \textbf{0.681} \\
\Xhline{3\arrayrulewidth}
\end{tabular}
\caption{Comparison studies of \pipelines \ on CMU-MOSI. Modeling cross-modal interactions using multistage fusion and attention weights are crucial in multimodal language analysis. \vspace{-4mm}}
\label{ablation}
\end{table}

To achieve a deeper understanding of the multistage fusion process, we study five research questions. (Q1): whether modeling cross-modal interactions across multiple stages is beneficial. (Q2): the effect of the number of stages $K$ during multistage fusion on performance. (Q3): the comparison between multistage and independent modeling of cross-modal interactions. (Q4): whether modeling cross-modal interactions are helpful. (Q5): whether attention weights from the \texttt{HIGHLIGHT} module are required for modeling cross-modal interactions. 

\noindent \textbf{Q1:} To study the effectiveness of the multistage fusion process, we test the baseline \pipelines-R$1$ which performs fusion in only one stage instead of across multiple stages. This model makes the strong assumption that all cross-modal interactions can be modeled during only one stage. From Table~\ref{length}, \pipelines-R$1$ underperforms as compared to \pipelines \ which performs multistage fusion.

\noindent \textbf{Q2:} We test baselines \pipelines-R$K$ which perform $K$ stages of fusion. From Table~\ref{length}, we observe that increasing the number of stages $K$ increases the model's capability to model cross-modal interactions up to a certain point ($K=3$) in our experiments. Further increases led to decreases in performance and we hypothesize this is due to overfitting on the dataset.

\noindent \textbf{Q3:} To compare multistage against independent modeling of cross-modal interactions, we pay close attention to the performance comparison with respect to MARN which models multiple cross-modal interactions all at once (see Table~\ref{ablation}). \pipelines \ shows improved performance, indicating that multistage fusion is both effective and efficient for human multimodal language modeling.

\noindent \textbf{Q4:} \pipelines \ (no \ours) represents a system of LSTHMs without the integration of $\mathbf{z}_t$ from the \ours \ to model cross-modal interactions. From Table~\ref{ablation}, \pipelines \ (no \ours) is outperformed by \pipelines, confirming that modeling cross-modal interactions is crucial in analyzing human multimodal language. 

\noindent \textbf{Q5:} \pipelines \ (no \texttt{HIGHLIGHT}) removes the \texttt{HIGHLIGHT} module from \ours \ during multistage fusion. From Table \ref{ablation}, \pipelines \ (no \texttt{HIGHLIGHT}) underperforms, indicating that highlighting multimodal representations using attention weights are important for modeling cross-modal interactions.

\subsection{Visualizations}

\begin{figure*}[t!]
\centering{
\includegraphics[width=0.9\linewidth]{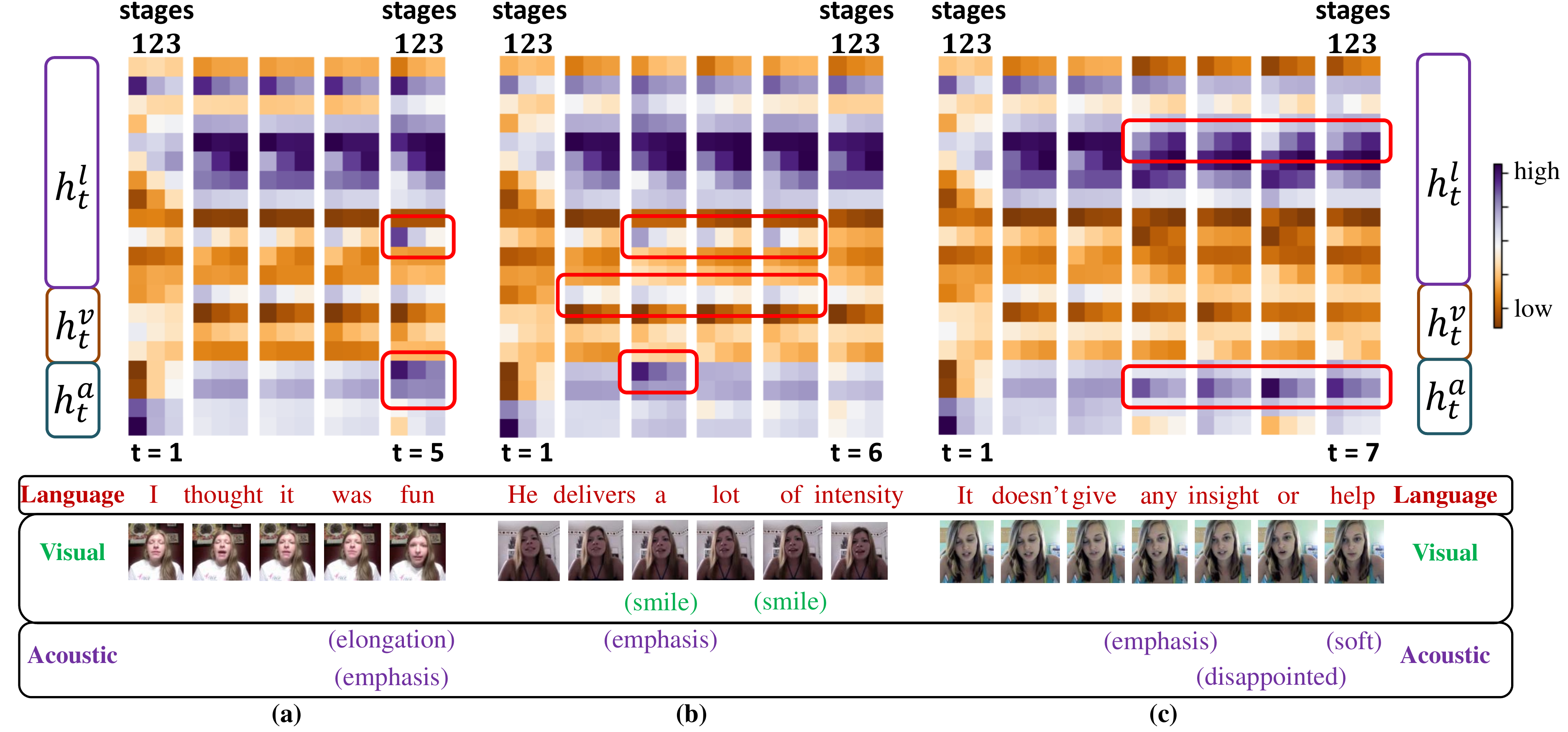}}
    \caption{Visualization of learned attention weights across stages 1,2 and 3 of the multistage fusion process and across time of the multimodal sequence. We observe that the attention weights are diverse and evolve across stages and time. In these three examples, the red boxes emphasize specific moments of interest. (a) Synchronized interactions: the positive word ``fun'' and the acoustic behaviors of emphasis and elongation ($t=5$) are synchronized in both attention weights for language and acoustic features. (b) Asynchronous trimodal interactions: the asynchronous presence of a smile ($t=2:5$) and emphasis ($t=3$) help to disambiguate the language modality. (c) Bimodal interactions: the interactions between the language and acoustic modalities are highlighted by alternating stages of fusion ($t=4:7$). \vspace{-4mm}}
    \label{fig:vis}
\end{figure*}


Using an attention assignment mechanism during the \texttt{HIGHLIGHT} process gives more interpretability to the model since it allows us to visualize the attended multimodal signals at each stage and time step (see Figure~\ref{fig:vis}). Using \pipelines \ trained on the CMU-MOSI dataset, we plot the attention weights across the multistage fusion process for three videos in CMU-MOSI. Based on these visualizations we first draw the following general observations on multistage fusion:

\noindent \textbf{Across stages:} Attention weights change their behaviors across the multiple stages of fusion. Some features are highlighted by earlier stages while other features are used in later stages. This supports our hypothesis that \pipelines \ learns to specialize in different stages of the fusion process.

\noindent \textbf{Across time:} Attention weights vary over time and adapt to the multimodal inputs. We observe that the attention weights are similar if the input contains no new information. As soon as new multimodal information comes in, the highlighting mechanism in \pipelines \ adapts to these new inputs.


\noindent \textbf{Priors:} Based on the distribution of attention weights, we observe that the language and acoustic modalities seem the most commonly highlighted. This represents a prior over the expression of sentiment in human multimodal language and is closely related to the strong connections between language and speech in human communication~\citep{Kuhl11850}.


\noindent \textbf{Inactivity:} Some attention coefficients are not active (always orange) throughout time. We hypothesize that these corresponding dimensions carry only intra-modal dynamics and are not involved in the formation of cross-modal interactions.

\subsection{Qualitative Analysis}

In addition to the general observations above, Figure~\ref{fig:vis} shows three examples where multistage fusion learns cross-modal representations across three different scenarios.

\noindent \textbf{Synchronized Interactions:} In Figure~\ref{fig:vis}(a), the language features are highlighted corresponding to the utterance of the word ``fun'' that is highly indicative of sentiment ($t=5$). This sudden change is also accompanied by a synchronized highlighting of the acoustic features. We also notice that the highlighting of the acoustic features lasts longer across the 3 stages since it may take multiple stages to interpret all the new acoustic behaviors (elongated tone of voice and phonological emphasis).

\noindent \textbf{Asynchronous Trimodal Interactions:} In Figure~\ref{fig:vis}(b), the language modality displays ambiguous sentiment: ``delivers a lot of intensity'' can be inferred as both positive or negative. We observe that the circled attention units in the visual and acoustic features correspond to the asynchronous presence of a smile ($t=2:5$) and phonological emphasis ($t=3$) respectively. These nonverbal behaviors resolve ambiguity in language and result in an overall display of positive sentiment. We further note the coupling of attention weights that highlight the language, visual and acoustic features across stages ($t=3:5$), further emphasizing the coordination of all three modalities during multistage fusion despite their asynchronous occurrences.

\noindent \textbf{Bimodal Interactions:} In Figure~\ref{fig:vis}(c), the language modality is better interpreted in the context of acoustic behaviors. The disappointed tone and soft voice provide the nonverbal information useful for sentiment inference. This example highlights the bimodal interactions ($t=4:7$) in alternating stages: the acoustic features are highlighted more in earlier stages while the language features are highlighted increasingly in later stages.



\section{Conclusion}
This paper proposed the \pipelinel \ (\pipelines) which decomposes the multimodal fusion problem into multiple stages, each focused on a subset of multimodal signals. Extensive experiments across three publicly-available datasets reveal that \pipelines \ is highly effective in modeling human multimodal language. In addition to achieving state-of-the-art performance on all datasets, our comparisons and visualizations reveal that the multiple stages coordinate to capture both synchronous and asynchronous multimodal interactions. In future work, we are interested in merging our model with memory-based fusion methods since they have complementary strengths as discussed in subsection~\ref{results}.


\section{Acknowledgements}

The authors thank Yao Chong Lim, Venkata Ramana Murthy Oruganti, Zhun Liu, Ying Shen, Volkan Cirik, and the anonymous reviewers for their constructive comments
on this paper.

\bibliography{emnlp2018}
\bibliographystyle{acl_natbib_nourl}

\title{Multimodal Language Analysis with Recurrent Multistage Fusion: \\ Supplementary Material}

\author{Paul Pu Liang$^1$, Ziyin Liu$^2$, Amir Zadeh$^2$, Louis-Philippe Morency$^2$ \\
$^1$Machine Learning Department, $^2$Language Technologies Institute \\
Carnegie Mellon University \\
\texttt { \{pliang,ziyinl,abagherz,morency\}@cs.cmu.edu} \\
}
\date{}

\maketitle

\section{Supplementary Material}

\subsection{Multimodal Features}

Here we present extra details on feature extraction for the language, visual and acoustic modalities.

\noindent \textbf{Language:} We used 300 dimensional Glove word embeddings trained on 840 billion tokens from the common crawl dataset \cite{pennington2014glove}. These word embeddings were used to embed a sequence of individual words from video segment transcripts into a sequence of word vectors that represent spoken text. 

\noindent \textbf{Visual:} The library Facet \cite{emotient} is used to extract a set of visual features including facial action units, facial landmarks, head pose, gaze tracking and HOG features \cite{zhu2006fast}. These visual features are extracted from the full video segment at 30Hz to form a sequence of facial gesture measures throughout time.

\noindent \textbf{Acoustic:} The software COVAREP \cite{degottex2014covarep} is used to extract acoustic features including 12 Mel-frequency cepstral coefficients, pitch tracking and voiced/unvoiced segmenting features \cite{drugman2011joint}, glottal source parameters \cite{childers1991vocal,drugman2012detection,alku1992glottal,alku1997parabolic,alku2002normalized}, peak slope parameters and maxima dispersion quotients \cite{kane2013wavelet}. These visual features are extracted from the full audio clip of each segment at 100Hz to form a sequence that represent variations in tone of voice over an audio segment.

\subsection{Multimodal Alignment}
We perform forced alignment using P2FA~\cite{P2FA} to obtain the exact utterance time-stamp of each word. This allows us to align the three modalities together. Since words are considered the basic units of language we use the interval duration of each word utterance as one time-step. We acquire the aligned video and audio features by computing the expectation of their modality feature values over the word utterance time interval~\cite{factorized}.

\subsection{Additional Results}

Here we record the complete set of results for all the baseline models across all the datasets, tasks and metrics. Table~\ref{mosi_supp} summarizes the complete results for sentiment analysis on the CMU-MOSI dataset. Table~\ref{iemocap_supp} presents the complete results for emotion recognition on the IEMOCAP dataset and Table~\ref{pom_supp} presents the complete results for personality traits prediction on the POM dataset. For experiments on the POM dataset we report additional results on MAE and correlation metrics for personality traits regression. We achieve improvements over state-of-the-art multi-view and dataset specific approaches across all these datasets, highlighting the \pipelines's capability in analyzing sentiment, emotions and speaker traits from human multimodal language.

\newcolumntype{K}[1]{>{\centering\arraybackslash}p{#1}}
\definecolor{gg}{RGB}{45,190,45}

\begin{table}[t!]
\fontsize{7.5}{10}\selectfont
\setlength\tabcolsep{1.3pt}
\begin{tabular}{l : *{16}{K{1.05cm}}}
\Xhline{3\arrayrulewidth}
Dataset & \multicolumn{5}{c}{\textbf{CMU-MOSI}} \\
Task & \multicolumn{5}{c}{Sentiment} \\
Metric       & A$^{2}$ & F1 & A$^{7}$ & MAE & Corr\\ 
\Xhline{0.5\arrayrulewidth}
Majority       & 50.2 &   50.1   &	 17.5	&     1.864 &  0.057  \\
RF         & 56.4 &  56.3    &  21.3 	&		-     	&  - \\ 
SVM-MD     & 71.6 &   72.3   & 26.5  	& 1.100     &  0.559 \\ 
THMM		& 50.7	& 45.4	& 17.8& - & -\\
SAL-CNN    & 73.0 &   -      &  -	&  -        	& -\\ 
C-MKL          & 72.3 &   72.0   &  30.2	&  -       		& -\\
EF-HCRF			& 65.3 & 65.4 & 24.6 & - & -\\
EF-LDHCRF		& 64.0 & 64.0 & 24.6 & - & -\\
MV-HCRF			& 44.8 & 27.7 & 22.6 & - & -\\
MV-LDHCRF		& 64.0 & 64.0 & 24.6 & - & -\\
CMV-HCRF		& 44.8 & 27.7 & 22.3 & - & -\\
CMV-LDHCRF		& 63.6 & 63.6 & 24.6 & - & -\\
EF-HSSHCRF		& 63.3 & 63.4 & 24.6 & - & -\\
MV-HSSHCRF		& 65.6 & 65.7 & 24.6 & - & -\\
DF              & 72.3 &   72.1   &  26.8 	& 1.143     &  0.518 \\
EF-LSTM         & 74.3 &   74.3   &  32.4 	& 1.023     &  0.622 \\
EF-SLSTM		& 72.7 & 72.8 & 29.3 & 1.081 & 0.600 \\
EF-BLSTM		& 72.0 & 72.0 & 28.9 & 1.080 & 0.577 \\
EF-SBLSTM		& 73.3 &   73.2   &  26.8   & 1.037 & 0.619 \\
MV-LSTM			& 73.9 &   74.0   &  33.2	& 1.019 &  0.601 \\
BC-LSTM         & 73.9 &   73.9   &  28.7	& 1.079 &  0.581 \\ 
TFN             & 74.6 &   74.5   &  28.7	& 1.040	& 0.587   \\ 
GME-LSTM(A) & 76.5 & 73.4 & - & {0.955} & - \\
MARN & 77.1	& 77.0 & 34.7 & 0.968  & 0.625 \\ 
MFN & {77.4}	 & {77.3}   & {34.1} 	&  {0.965}    	& {0.632} \\
\Xhline{0.5\arrayrulewidth}
{\pipelines}      		& \textbf{78.4} & \textbf{78.0} & \textbf{38.3} & \textbf{0.922} & \textbf{0.681} \\
\Xhline{0.5\arrayrulewidth}
$\Delta_{SOTA}$	& \textcolor{gg}{$\uparrow $ \textbf{1.0}} & \textcolor{gg}{$\uparrow $ \textbf{0.7}} & \textcolor{gg}{$\uparrow $ \textbf{3.6}} & \textcolor{gg}{$\downarrow $ \textbf{0.033}} & \textcolor{gg}{$\uparrow $ \textbf{0.049}} \\ 
\Xhline{0.5\arrayrulewidth}
Human  & 85.7 &   87.5   &  53.9	&    0.710  &  0.820  \\ \Xhline{3\arrayrulewidth}
\end{tabular}
\caption{Sentiment prediction results on CMU-MOSI test set. The best results are highlighted in bold and $\Delta_{SOTA}$ shows the change in performance over previous state of the art (SOTA). Improvements are highlighted in green. The \pipelines \ significantly outperforms the current SOTA across all evaluation metrics.}
\label{mosi_supp}
\end{table}
\definecolor{gg}{RGB}{45,190,45}

\begin{table}[tb]
\fontsize{7.5}{10}\selectfont
\setlength\tabcolsep{1.3pt}
\begin{tabular}{l : *{16}{K{0.63cm}}}
\Xhline{3\arrayrulewidth}
Dataset & \multicolumn{8}{c}{\textbf{IEMOCAP Emotions}} \\
Task & \multicolumn{2}{c}{Happy} & \multicolumn{2}{c}{Sad} & \multicolumn{2}{c}{Angry} & \multicolumn{2}{c}{Neutral} \\
Metric        	& A$^{2}$ & F1 & A$^{2}$ & F1 & A$^{2}$ & F1 & A$^{2}$ & F1 \\ 
\Xhline{0.5\arrayrulewidth}
Majority		& 85.6 & 79.0 & 79.4 & 70.3 & 75.8 & 65.4 & 59.1 & 44.0 \\
SVM     		& 86.1 & 81.5 & 81.1 & 78.8 & 82.5 & 82.4 & 65.2 & 64.9 \\
RF     			& 85.5 & 80.7 & 80.1 & 76.5 & 81.9 & 82.0 & 63.2 & 57.3 \\
THMM			&  85.6 & 79.2 & 79.5 & 79.8 & 79.3 & 73.0 & 58.6 & 46.4 \\
EF-HCRF			& 85.7 & 79.2 & 79.4 & 70.3 & 75.8 & 65.4 & 59.1 & 44.0 \\
EF-LDHCRF		& 85.8 & 79.5 & 79.4 & 70.3 & 75.8 & 65.4 & 59.1 & 44.0 \\
MV-HCRF			& 15.0 & 4.9  & 79.4 & 70.3 & 24.2 & 9.4  & 59.1 & 44.0 \\
MV-LDHCRF		& 85.7 & 79.2 & 79.4 & 70.3 & 75.8 & 65.4 & 59.1 & 44.0 \\
CMV-HCRF		& 14.4 & 3.6  & 79.4 & 70.3 & 24.2 & 9.4  & 59.1 & 44.0 \\
CMV-LDHCRF		& 85.8 & 79.5 & 79.4 & 70.3 & 75.8 & 65.4 & 59.1 & 44.0 \\
EF-HSSHCRF		& 85.8 & 79.5 & 79.4 & 70.3 & 75.8 & 65.4 & 59.1 & 44.0 \\
MV-HSSHCRF		& 85.8 & 79.5 & 79.4 & 70.3 & 75.8 & 65.4 & 59.1 & 44.0 \\
DF   			& 86.0 & 81.0 & 81.8 & 81.2 & 75.8 & 65.4 & 59.1 & 44.0 \\
EF-LSTM   		& 85.2 & 83.3 & 82.1 & 81.1 & 84.5 & 84.3 & 68.2 & 67.1 \\
EF-SLSTM		& 85.6 & 79.0 & 80.7 & 80.2 & 82.8 & 82.2 & 68.8 & 68.5 \\
EF-BLSTM 		& 85.0 & 83.7 & 81.8 & 81.6 & 84.2 & 83.3 & 67.1 & 66.6 \\
EF-SBLSTM 		& 86.0 & 84.2 & 80.2 & 80.5 & \textbf{85.2} & {84.5} & 67.8 & 67.1 \\
MV-LSTM   		& 85.9 & 81.3 & 80.4 & 74.0 & 85.1 & 84.3 & 67.0 & 66.7 \\
BC-LSTM    		& 84.9 & 81.7 & 83.2 & 81.7 & 83.5 & 84.2 & 67.5 & 64.1 \\
TFN      		& 84.8 & 83.6 & 83.4 & {82.8} & 83.4 & 84.2 & 67.5 & 65.4 \\ 
MARN			& 86.7 & 83.6 & 82.0 & 81.2 & 84.6 & 84.2 & 66.8 & 65.9 \\
MFN				& 86.5 & 84.0 & 83.5 & 82.1 & 85.0 & 83.7 & \textbf{69.6} & \textbf{69.2} \\
\Xhline{0.5\arrayrulewidth}
{\pipelines}      		& \textbf{87.5} & \textbf{85.8} & \textbf{83.8} & \textbf{82.9} & {85.1} & \textbf{84.6} & {69.5} & {69.1} \\ 
\Xhline{0.5\arrayrulewidth}
$\Delta_{SOTA}$	& \textcolor{gg}{$\uparrow $ \textbf{0.8}} & \textcolor{gg}{$\uparrow $ \textbf{1.6}} & \textcolor{gg}{$\uparrow $ \textbf{0.3}} & \textcolor{gg}{$\uparrow $ \textbf{0.1}} & \textcolor{gray}{$\downarrow $ \textbf{0.1}} & \textcolor{gg}{$\uparrow $ \textbf{0.1}} & \textcolor{gray}{$\downarrow $ \textbf{0.1}} & \textcolor{gray}{$\downarrow $ \textbf{0.1}} \\ 
\Xhline{3\arrayrulewidth}
\end{tabular}
\caption{Emotion recognition results on IEMOCAP test set. The best results are highlighted in bold and $\Delta_{SOTA}$ shows the change in performance over previous SOTA. Improvements are highlighted in green. The \pipelines \ achieves state-of-the-art or competitive performance across all evaluation metrics.}
\label{iemocap_supp}
\end{table}
\newcolumntype{K}[1]{>{\centering\arraybackslash}p{#1}}
\definecolor{gg}{RGB}{45,190,45}

\begin{table*}[!htbp]
\fontsize{7}{10}\selectfont
\setlength\tabcolsep{1.3pt}
\begin{tabular}{l : *{16}{K{1.11cm}}}
\Xhline{3\arrayrulewidth}
Dataset & \multicolumn{16}{c}{\textbf{POM Speaker Personality Traits}} \\
Task & \multicolumn{1}{c}{Con} & \multicolumn{1}{c}{Pas} & \multicolumn{1}{c}{Voi} & \multicolumn{1}{c}{Cre} & \multicolumn{1}{c}{Viv} & \multicolumn{1}{c}{Exp}& \multicolumn{1}{c}{Res} & \multicolumn{1}{c}{Rel} & \multicolumn{1}{c}{Tho} & \multicolumn{1}{c}{Ner} & \multicolumn{1}{c}{Per} & \multicolumn{1}{c}{Hum}\\
Metric & A$^7$ & A$^7$ & A$^7$ & A$^7$ &  A$^7$ & A$^7$  & A$^5$  & A$^5$ & A$^5$ & A$^5$ & A$^7$ & A$^5$  \\ 
\Xhline{0.5\arrayrulewidth}
Majority		& 19.2&20.2&30.5&21.7&25.6&26.1&29.6&39.4&31.0&24.1&20.7&6.9 \\
SVM     		&  20.6 & 20.7 & 32.0 & 25.1 & 29.1 & 34.0 & 49.8 & 42.9 & 39.9 & 41.4 & 28.1 & 36.0 \\
RF     			&  26.6&27.1&29.6&23.2&23.6&26.6&34.0&40.9&37.4&36.0&25.6&40.4  \\
THMM    		& 24.1&15.3&19.2&27.6&26.1&18.7&22.7&31.5&30.0&27.1&17.2&24.6 \\
DF   & 25.6&24.1&33.0&26.1&32.0&26.6&30.0&50.2&37.9&42.4&26.6&34.5\\
EF-LSTM   & 20.7&27.6&31.5&25.1&31.0&25.1&30.0&48.3&42.4&40.4&25.6&36.0\\
EF-SLSTM	& 22.2&28.6&30.5&27.1&32.0&27.6&32.5&46.8&39.9&41.9&22.7&35.0 \\
EF-BLSTM & 25.1&26.1&34.0&29.6&31.0&25.6&30.0&46.3&41.9&42.9&25.6&39.4\\
EF-SBLSTM & 23.2&30.5&29.1&27.6&32.5&31.0&33.5&47.8&39.4&44.8&25.6&38.9\\
MV-LSTM 		& 25.6&28.6&28.1&25.6&32.5&29.6&33.0&50.7&37.9&42.4&26.1&38.9\\
BC-LSTM    		& 26.6&26.6&31.0&27.6&36.5&30.5&33.0&47.3&45.8&36.0&27.1&36.5\\ 
TFN      		& 24.1&31.0&31.5&24.6&25.6&27.6&30.5&35.5&33.0&42.4&27.6&33.0\\ 
MARN		& 29.1&33.0&-&31.5&-&-&36.9&52.2&-&47.3&31.0&44.8 \\
MFN			& {34.5}&{35.5}&{37.4}&{34.5}&{36.9}&{36.0}&{38.4}&{53.2}&{47.3}&{47.8}&{34.0}&\textbf{47.3} \\
\Xhline{0.5\arrayrulewidth}
{\pipelines }  &  \textbf{37.4} & \textbf{38.4} &\textbf{37.4}  &\textbf{37.4} &\textbf{38.9} &\textbf{38.9}  &\textbf{39.4} &\textbf{53.7} &\textbf{48.3} &\textbf{48.3} &\textbf{35.0} &{46.8}\\
\Xhline{0.5\arrayrulewidth}
$\Delta_{SOTA}$	& \textcolor{gg}{$\uparrow $ \textbf{2.9}} & \textcolor{gg}{$\uparrow $ \textbf{2.9}} & \textcolor{gg}{$\uparrow $ \textbf{0.0}} & \textcolor{gg}{$\uparrow $ \textbf{2.9}} & \textcolor{gg}{$\uparrow $ \textbf{2.0}} & \textcolor{gg}{$\uparrow $ \textbf{2.9}}  & \textcolor{gg}{$\uparrow $ \textbf{1.0}} & \textcolor{gg}{$\uparrow $ \textbf{0.5}} & \textcolor{gg}{$\uparrow $ \textbf{1.0}} & \textcolor{gg}{$\uparrow $ \textbf{0.5}} & \textcolor{gg}{$\uparrow $ \textbf{1.0}} & \textcolor{gray}{$\uparrow $ \textbf{0.5}} \\ 
\Xhline{3\arrayrulewidth}
\end{tabular}
\end{table*}

\begin{table*}[!htbp]
\fontsize{7}{10}\selectfont
\setlength\tabcolsep{1.3pt}
\begin{tabular}{l : *{16}{K{1.11cm}}}
\Xhline{3\arrayrulewidth}
Dataset & \multicolumn{16}{c}{\textbf{POM Speaker Personality Traits}} \\
Task & \multicolumn{1}{c}{Con} & \multicolumn{1}{c}{Pas} & \multicolumn{1}{c}{Voi}  & \multicolumn{1}{c}{Cre} & \multicolumn{1}{c}{Viv} & \multicolumn{1}{c}{Exp} & \multicolumn{1}{c}{Res} & \multicolumn{1}{c}{Rel} & \multicolumn{1}{c}{Tho} & \multicolumn{1}{c}{Ner} & \multicolumn{1}{c}{Per} & \multicolumn{1}{c}{Hum}\\
Metric & \multicolumn{16}{c}{MAE} \\ 
\Xhline{0.5\arrayrulewidth}
Majority & 1.483&1.268&1.089&1.260&1.158&1.164&1.166&0.753&0.939&1.181&1.328&1.774 \\
SVM	& 1.071&1.159&0.938&1.036&1.043&1.031&0.877&0.594&0.728&0.714&1.110&0.801\\
DF		& 1.033&1.083&0.899&1.039&0.997&1.027&0.884&0.591&0.732&0.695&1.086&0.768 \\
EF-LSTM			&  1.035&1.067&0.911&1.022&0.981&0.990&0.880&0.594&0.712&0.706&1.084&0.768\\
EF-SLSTM & 1.047&1.069&0.924&1.040&0.990&1.005&0.872&0.597&0.726&0.686&1.098&0.787 \\
EF-BLSTM & 1.103&1.105&0.975&1.053&1.018&1.069&0.882&0.607&0.762&0.705&1.085&0.762\\
EF-SBLSTM &  1.062&1.055&0.926&1.027&1.020&0.991&0.877&0.594&0.746&0.697&1.102&0.779\\
MV-LSTM	& 	1.029&1.098&0.971&1.082&0.976&1.012&0.877&0.625&0.792&0.687&1.163&0.770 \\
BC-LSTM		& 1.016&1.008&0.914&0.942&\textbf{0.905}&0.906&0.888&0.630&0.680&0.705&1.025&0.767	 \\
TFN  & 1.049&1.104&0.927&1.058&1.000&1.029&0.900&0.621&0.743&0.727&1.132&0.770\\
MFN			&
{0.952}&{0.993}&{0.882}&\textbf{0.903}&0.908&\textbf{0.886}&{0.821}&{0.566}&\textbf{0.665}&{0.654}&\textbf{0.981}&\textbf{0.727} \\  
\Xhline{0.5\arrayrulewidth}
{\pipelines } 	& 
\textbf{0.933}&\textbf{0.938}&\textbf{0.868}&{0.967}&{0.914}&{0.887}&\textbf{0.781}&\textbf{0.566}&{0.666}&\textbf{0.640}&{1.014}&{0.728} \\ \Xhline{0.5\arrayrulewidth}
$\Delta_{SOTA}$	& \textcolor{gg}{$\downarrow $ \textbf{0.019}} & \textcolor{gg}{$\downarrow $ \textbf{0.055}} & \textcolor{gg}{$\downarrow $ \textbf{0.014}}  & \textcolor{gray}{$\downarrow $ \textbf{0.064}}  & \textcolor{gray}{$\downarrow $ \textbf{0.001}} & \textcolor{gg}{$\downarrow $ \textbf{0.020}} &\textcolor{gg}{$\downarrow $ \textbf{0.040}}  & \textcolor{gg}{$\downarrow $ \textbf{0.0}} & \textcolor{gray}{$\uparrow $ \textbf{0.001}} & \textcolor{gg}{$\downarrow $ \textbf{0.014}} & \textcolor{gray}{$\uparrow $ \textbf{0.033}} & \textcolor{gray}{$\downarrow $ \textbf{0.001}}\\ 
\Xhline{3\arrayrulewidth}
\end{tabular}
\label{overall-3}
\end{table*}

\begin{table*}[!htbp]
\fontsize{7}{10}\selectfont
\setlength\tabcolsep{1.3pt}
\begin{tabular}{l : *{16}{K{1.11cm}}}
\Xhline{3\arrayrulewidth}
Dataset & \multicolumn{16}{c}{\textbf{POM Speaker Personality Traits}} \\
Task & \multicolumn{1}{c}{Con} & \multicolumn{1}{c}{Pas} & \multicolumn{1}{c}{Voi} & \multicolumn{1}{c}{Cre} & \multicolumn{1}{c}{Viv} & \multicolumn{1}{c}{Exp} & \multicolumn{1}{c}{Res} & \multicolumn{1}{c}{Rel} & \multicolumn{1}{c}{Tho} & \multicolumn{1}{c}{Ner} & \multicolumn{1}{c}{Per} & \multicolumn{1}{c}{Hum}\\
Metric  & \multicolumn{16}{c}{$r$} \\ 
\Xhline{0.5\arrayrulewidth}
Majority & -0.041&-0.029&-0.104&-0.122&-0.044&-0.065&0.006&-0.024&-0.130&0.097&-0.127&-0.069 \\
SVM	& 0.063&0.086&-0.004&0.113&0.076&0.134&0.166&0.104&0.134&0.068&0.064&0.147\\
DF	& 0.240&0.273&0.017&0.112&0.173&0.118&0.148&0.019&0.041&0.136&0.168&0.259 \\
EF-LSTM			& 0.200&0.302&0.031&0.170&0.244&0.265&0.142&0.083&0.260&0.105&0.217&0.227\\
EF-SLSTM & 0.221&0.327&0.042&0.177&0.239&0.268&0.204&0.092&0.252&0.159&0.218&0.196 \\
EF-BLSTM & 
0.162&0.289&0.034&0.191&0.279&0.274&0.184&0.093&0.245&0.166&0.243&0.272\\
EF-SBLSTM  & 0.174&0.310&0.021&0.170&0.224&0.261&0.155&0.097&0.215&0.121&0.216&0.171\\
MV-LSTM	& 0.358&0.416&0.131&0.280&0.347&0.323&0.295&0.119&0.284&0.258&0.239&0.317 \\
BC-LSTM		& 0.359&0.425&0.081&0.358&0.417&0.450&0.293&0.075&0.363&0.184&0.344&0.319\\
TFN &  0.089&0.201&0.030&0.124&0.204&0.171&-0.051&0.114&0.048&-0.002&0.106&0.213 \\ 
MFN  &  {0.395}&{0.428}&{0.193}&\textbf{0.367}&\textbf{0.431}&{0.452}&{0.333}&{0.255}&\textbf{0.381}&{0.318}&\textbf{0.377}&\textbf{0.386}\\ 
\Xhline{0.5\arrayrulewidth} 
{\pipelines }  &  \textbf{0.441}&\textbf{0.502}&\textbf{0.247}&{0.355}&{0.414}&\textbf{0.470}&\textbf{0.446}&\textbf{0.291}&{0.376}&\textbf{0.379}&{0.327}&{0.346}\\ 
\Xhline{0.5\arrayrulewidth}
$\Delta_{SOTA}$	& \textcolor{gg}{$\uparrow $ \textbf{0.046}} & \textcolor{gg}{$\uparrow $ \textbf{0.074}} & \textcolor{gg}{$\uparrow $ \textbf{0.054}} & \textcolor{gray}{$\downarrow $ \textbf{0.012}} & \textcolor{gray}{$\downarrow $ \textbf{0.017}} & \textcolor{gg}{$\downarrow $ \textbf{0.018}}  & \textcolor{gg}{$\uparrow $ \textbf{0.113}}  & \textcolor{gg}{$\uparrow $ \textbf{0.036}} & \textcolor{gray}{$\downarrow $ \textbf{0.005}} & \textcolor{gg}{$\uparrow $ \textbf{0.061}}& \textcolor{gray}{$\downarrow $ \textbf{0.050}} & \textcolor{gray}{$\downarrow $ \textbf{0.040}} \\ \Xhline{3\arrayrulewidth}
\end{tabular}
\caption{Results for personality trait recognition on the POM dataset. The best results are highlighted in bold and $\Delta_{SOTA}$ shows the change in performance over previous state of the art. Improvements are highlighted in green. The \pipelines \ achieves state-of-the-art or competitive performance across all evaluation metrics.}
\label{pom_supp}
\end{table*}


\end{document}